\documentclass[preprint,12pt]{elsarticle}

\usepackage{albi_pack}
\usepackage{paralist}
\usepackage{changepage}

\journal{Engineering Applications of Artificial Intelligence}

\begin{document}

\begin{frontmatter}

\title{CONFIDERAI: a novel CONFormal Interpretable-by-Design score function for Explainable and Reliable Artificial Intelligence}

\author[inst1,inst2]{Sara Narteni\corref{c1}$^\dagger$}
\cortext[c1]{\textit{Corresponding Author}\\Corso F.M. Perrone 24, 16152, Genoa, Italy\\ \textit{Email Address: } saranarteni@cnr.it (Sara Narteni)\\$^\dagger$ These authors contributed equally to the development of the article.}
\author[inst1]{Alberto Carlevaro$^\dagger$}
\author[inst1]{Fabrizio Dabbene}
\author[inst1,inst3]{Marco Muselli}
\author[inst1]{Maurizio Mongelli}
\affiliation[inst1]{organization={Institute of Electronics, Information Engineering and Telecommunications - National Research Council of Italy (CNR-IEIIT)}}
\affiliation[inst2]{organization={Politecnico di Torino - Department of Control and Computer Engineering (DAUIN)},addressline={Corso Duca degli Abruzzi 24}, city={Torino},postcode={10129}, state={Italy}}
\affiliation[inst3]{organization={Rulex Innovation Labs},
            addressline={Via Felice Romani 9}, 
            city={Genoa},
            postcode={16122}, 
            state={Italy}}

\begin{abstract}
Everyday life is increasingly influenced by artificial intelligence, and there is no question that machine learning algorithms must be designed to be reliable and trustworthy for everyone. Specifically, computer scientists consider an artificial intelligence system safe and trustworthy if it fulfills five pillars: explainability, robustness, transparency, fairness, and privacy. In addition to these five, we propose a sixth fundamental aspect: conformity, that is, the probabilistic assurance that the system will behave as the machine learner expects. In this paper, we present a methodology to link conformal prediction with explainable machine learning by defining a new score function for rule-based classifiers that leverages rules predictive ability, the geometrical position of points within rules boundaries and the overlaps among rules as well, thanks to the definition of a geometrical rule similarity term. 
Furthermore, we address the problem of defining regions in the feature space where conformal guarantees are satisfied, by exploiting the definition of conformal critical set and showing how this set can be used to achieve new rules with improved performance on the target class.
The overall methodology is tested with promising results on several datasets of real-world interest, such as domain name server tunneling detection or cardiovascular disease prediction.
\end{abstract}

\begin{highlights}
\item Performance guarantees for rule-based classification models
\item Novel score function accounting for rule overlaps via geometrical rule similarity
\item Conformal prediction-guided tuning of rules via conformal critical set
\end{highlights}


\end{frontmatter}

\section{Introduction}\label{sec:intro}

Trustworthy Artificial Intelligence (AI) is an umbrella-term that gained increasing importance in recent years to establish the requirements of real-world AI systems for their proper design, development and deployment. Among its principles, \textit{explainability} (or transparency) and \textit{technical robustness and safety} (or reliability) have an essential role \cite{trustworthyaiguidelines}.
Explainability allows each actor involved to understand the reasoning behind any machine learning (ML) decision. There is a plethora of techniques to achieve explainability today, falling under the \textit{eXplainable AI (XAI)} research theme \cite{ali2023explainable}. At a high level, the main categorization of XAI distinguishes post-hoc explanations of black box models and transparent-by-design techniques~\cite{minh2022explainable}. The latter category includes rule-based models, where predictions are characterized by easy-to-understand decision rules (often expressed in the \textit{if-then} format).

Nevertheless, even though very helpful thanks to their native interpretability, rule-based models alone are not enough to ensure a correct performance of the model. Therefore, many approaches have been proposed so far to guarantee the safety of ML models \cite{Neto22,dey2021multilayered}, also relying on (or devoted to) XAI methodologies\cite{marques2022delivering,hurault2023certified,angelino2017learning,narteni2022sensitivity}. Among them, conformal prediction (CP) stands out with its solid mathematical foundation, that allows to generate prediction sets with predefined probabilistic guarantees for any ML model \cite{tutorial_CP}. 
However, as  discussed in Section~\ref{sec:relworks}, we were not able to find in the current literature works specifically focusing on CP for transparent-by-design XAI models. Motivated by this observation, in this paper  we investigate this topic, by proposing an innovative score function that enables CP for rule-based models.
On the other hand, various studies exploit CP to perform false positives or negatives control. For example, \cite{limitedFP} proposes a multi-label conformal prediction approach in which the false positive rate is probabilistically controlled by requiring prediction sets to eliminate non-conformal points. \cite{angelopoulos2022conformal} inserts the performance control directly on the expected value of any loss function and \cite{luo2022sample} introduces the concept of safety score that warns the system when a predefined level of error is reached. 
In our work, which is the extension of our preliminary work on this topic \cite{copa2023}, we will pursue the same objective by leveraging on \cite{carlevaro2024conformal}, i.e., defining a \textit{conformal critical set} in which the performance on a target class is guaranteed by the score function of the conformal prediction itself. 



\subsection{Contribution}\label{subsec:contr}
Based on the above considerations, in our viewpoint, combining CP framework with XAI is essential in the direction of trustworthy AI.
However, this topic is little explored in current literature, hence our paper attempts to address such a research gap through the following contributions:
\begin{itemize}
    \item We design and develop CONFIDERAI, a new score function to build conformal predictors for rule-based classifiers, by leveraging the combination of the global performance properties of decision rules (i.e., their covering and error) and the geometrical position of the points inside rule boundaries, as well as possible overlaps among rules.
    \item We exploit the concept of \emph{conformal critical set (CCS)}, i.e., the set of target points for which CONFIDERAI indicates high probabilistic guarantees of the underlying model, to improve the performance of the original rules. In particular, we show how this set leads to a new  labeling of the data useful to achieve a new set of rules with improved precision with respect to the original ones, still maintaining the explainability of the final classifier.
\end{itemize}

The remaining of the paper is structured as follows: Section~\ref{sec:relworks} reports existing approaches concerning CP and XAI models, with particular focus on rule-based ones; Section~\ref{sec:CSS} introduces the fundamentals of conformal prediction framework and provides the mathematical definition of the \emph{Conformal Critical Set}; Section~\ref{sec:RBC} describes our core contribution, the CONFIDERAI score function, along with simple toy examples to provide the reader with a visual intuition on how the score works; Section \ref{sec:CLLM} describes the fundamental properties of the specific rule-based model adopted in our case studies, i.e., the Logic Learning Machine (LLM); Section \ref{sec:Ex} reports and discusses the application of the proposed approach on relevant benchmarks and real-world datasets; finally, Section \ref{sec:conclusion} concludes the paper.

\section{Related Works}\label{sec:relworks}

As anticipated, conformal prediction for XAI models is still little investigated in research. 
Authors in \cite{cprandomforest} proposed the application of conformal predictors for evaluating the confidence of tree ensemble models, such as random forests. Also, \cite{wang2014reliable} studied CP for random forests in a multi-label classification scenario. Standard inductive conformal prediction (ICP) for classification through decision tree models is investigated in  \cite{cpusingdt2013}, and the same authors \cite{johansson2014rule} present such a framework to perform rule extraction with guarantees, either considering rule extraction \emph{for} opaque
models or rule extraction \emph{using} opaque models. Recently, in \cite{johansson2018interpretable,johansson2022rule}, the same approaches were extended to regression tasks. In all cases, authors define a conformity function based on the margin between true and predicted target probability estimates.


Even more recently, \cite{abdelqader2023interpretable} introduced a conformal decision rule learning algorithm, where rule generation and Mondrian conformal prediction \cite{vovk2003mondrian} processes are combined together, by devising a separate ICP on each rule at hand. Given unseen test samples satisfying a generic rule of any rule-based model, conformal decision rules provide a point prediction according to the consequent of the rule and a prediction set based on the results of conformal prediction for that rule. This approach assumes that the rules impose a partition of the training data into disjoint sets, which is a limitation when considering overlapping rules. Moreover, it adopts a canonical nonconformity measure based on nearest-neighbors within each rule, while our method considers distances of points with respect to the rules boundaries themselves.
Another study \cite{hullermeier2020conformal} involves the combination of rule-based models and CP frameworks for multi-label classification, by defining a conformity score that, for any candidate point, depends on the predictive quality of the top-performing rule. However, compared to our work, this score does not take into account the different position of instances within rule geometrical shape.
To the best of our knowledge, no previous study of these type address score functions and quantile, tailored for rule-based models (Sec. \ref{sec:CSS}-\ref{sec:RBC} here).

\section{Conformal Predictions and Critical Sets}
\label{sec:CSS}
Theory behind CP is substantially based on two, exchangeable, approaches: either $i)$ (non)-conformity measure and p-value as in \cite{tutorial_CP} or $ii)$ score functions and quantile as in \cite{gentle_CP}. As already pointed out, these two methodologies are actually the same but, in our opinion, the use of score functions and quantile is of more efficiency to understand properly the potential of XAI in CP, since score function links directly the conformal sets with the model. Moreover, we exploit the concept of \emph{conformal critical set} \cite{carlevaro2024conformal}, that allows to insert CP in a more safety-based context. As a matter of fact, our aim is to define a subset of the input feature space in which probabilistic guarantees can be provided to the ML model and to exploit explainable techniques to make it a fully trustworthy and easy-to-understand model.
\\

Let $\mathcal{X}$ be a measurable feature space and $\mathcal{Y}$ be the output space. We here consider the case of \textit{binary} classification, with   $\mathcal{Y}=\{0,+1\}$. Note that this choice of labels is without loss of generality, since any binary classifier can be converted to these labels.  Conformal prediction states that, for any machine learning model $\hat{f}(\x)_y: \mathcal{X} \longrightarrow \mathcal{Y}$, it is possible to define a \emph{score function} $s: \mathcal{X}\times\mathcal{Y} \longrightarrow \R$, which depends in some suitable way from the model:

\begin{equation*}
    s(\x,y) \sim \hat{f}(\x)_y.
\end{equation*}

\noindent The score function should be designed so that larger scores encode worse agreement between point $\x$ and label $y$.\\

In our context, we assume that the label $+1$ denotes the target class $S$, which is to be interpreted as the presence of \textit{critical situations} in the system.  The label $0$ refers to the non-target class instead, which denotes the absence of such conditions.

\begin{remark}[On the meaning of critical points]
Note that the meaning of the term ``critical'' is  context-dependent. For example, in a situation in which safety is of paramount importance, i.e.\ for instance in the case of collision avoidance, one may be interested in finding regions of the feature space where collision is avoided with high probability. In this case, the terms ``critical'' 
and ``safe" may be seen as synonyms\footnote{In this case, the label $+1$ will denote no collision, and $0$ will correspond to collision.}.
On the other hand, there are cases in which one would like to report only critical cases  when the probability of failure is very high, so to avoid false alarms. In this case, the class $+1$ would correspond to the ``failure" case.
\end{remark}

\noindent On the basis of the score function, a \emph{prediction set} at \emph{level of confidence} $1-\varepsilon$, $\varepsilon\in (0,1)$, can be defined for any $\x\in\mathcal{X}$:
\begin{equation}
    \mathcal{C}_\varepsilon(\x) = \{y\mid s(\x,y) \le s_{\varepsilon}\}\in2^{\mathcal{Y}},
    \label{eq:predictionset}
\end{equation}
\noindent where $s_{\varepsilon}$ is the $\ceil{(n_c+1)(1-\varepsilon)}/n_c$ \emph{quantile} of the score values computed on a \emph{calibration set} $\mathcal{Z}_c \doteq \left\{(\x_i,y_i)\right\}_{i=1}^{n_c},$ of size $n_c$. The prediction set guarantees the \textit{marginal coverage} property
\begin{equation}
   1-\varepsilon \le \Pr\{ y \in  \mathcal{C}_\varepsilon(\x) \}\le 1-\varepsilon + \frac{1}{n_c+1},
   \label{eq:margibal coverage}
\end{equation}
where ``marginal" means that the probability is averaged over the randomness of the calibration set. 

\noindent Keeping in mind all the above considerations on how to properly set a conformal prediction, we define the \emph{conformal critical set} (CCS) at \emph{confidence level} $1-\varepsilon$ the subset $\mathcal{S}_\varepsilon\subseteq\mathcal{X}$ as follows:
\begin{equation}
    \begin{split}
    \mathcal{S}_\varepsilon = \biggl\{\x\mid s(\x,+1)\le s_\varepsilon, \ s(\x,0)>s_\varepsilon\biggr\}.
    \end{split}
    \label{eq:css}
\end{equation}
\noindent In words, the CCS is a subset of the input space where the prediction set is composed by only unsafe points $(\x,+1)$.
%
%
This means that the model $\hat{f}$ is likely to make safe predictions for inputs in $\mathcal{S}_\varepsilon$ with a specified level of error $\varepsilon$.

\section{Rule-Based Conformity}\label{sec:RBC}

In the conformal prediction framework, as stated in Section \ref{sec:CSS}, any score function value $s(\x,y)$ is higher for any label $y$ that is less likely to be the correct prediction for the considered point $\x$. 
In this work, we aim at designing a new score function suitable for rule-based machine learning models.

\subsection{Rule-based classifiers notation}\label{subsec:RBnotation}

Before going into the design details, let us briefly describe the main characteristics and notation of rule-based models.

Let us consider an input dataset $\mathcal{T} = \{(\x_j, y_j)\}_{j=1}^N \in \mathcal{X}\times \mathcal{Y}, \text{ with } \x = (x_1,x_2, \dots,x_D) \in \mathbb{R}^{D} \text{ and } y \in \{0,1\}$. Also, assume a bounded feature space, i.e., it holds that $L_i\leq x_i \leq U_i,\; \forall i = 1, \ldots, D$.

A rule-based binary classifier $g: \mathcal{X}\rightarrow \mathcal{Y}$ is expressed by a set of decision rules $\mathcal{R} = \{r_k\}_{k=1}^{M_r}$ in the following form: \textcolor{blue}{\textbf{if}} \textit{premise} \textcolor{blue}{\textbf{then}} \textit{consequence}. 

The \textit{premise} of a rule $r_k$ is the logical conjunction ($\land$) of conditions on the input features, and is associated to a hyperrectangle $\mathcal{H}_{r_k}$ in the feature space, defined as:
\begin{equation}
    \mathcal{H}_{r_k} \doteq  \bigwedge_{i=1}^{D} c_{i_k},
    \label{eq:hyperrect}
\end{equation}
where  $c_{i_k}$ is an interval of the form $l_{i_k} \leq x_i \leq u_{i_k}$, being $l_{i_k} \geq L_i$ and $u_{i_k} \leq U_i$. 
It is frequent that some of the variables $x_i$ of the feature space are not explicitly expressed in the rules. However, this does not prevent us to consider (\ref{eq:hyperrect}), by setting  $L_i \leq x_i \leq U_i$ for such variables.
Moreover, a measure of the volume for $\mathcal{H}_{r_k}$ can be calculated as:
\begin{equation}
    \mathcal{V}_{\mathcal{H}_{r_k}} = \prod_{i=1}^{D} |u_{i_k}-l_{i_k}|.
    \label{eq:volumehyperrect}
\end{equation}

The \textit{consequence} of rule $r_k$ expresses the output class $\hat{y}_k \in \mathcal{Y}$ predicted by the decision rule.

\smallskip

Another useful concept in rule-based learning is the notion of \emph{rule relevance}, assigning to each rule a value in the [0,1] range which resembles its predictive ability. 
Specifically, it is computed by combining the covering $C(r_k)$ and error $E(r_k)$ metrics (commonly known as True Positive Rate and False Positive Rate of the rule, respectively), defined as follows:
\begin{equation}
    C(r_k) = \frac{TP(r_k)}{TP(r_k) + FN(r_k)}
\label{eq:covering}
\end{equation}
\begin{equation}
    E(r_k)=\frac{FP(r_k)}{TN(r_k) + FP(r_k)}
\label{eq:error}
\end{equation}
Given a set of points $\{(\x_j, y_j)\}$, $TP(r_k)$ and $FP(r_k)$ are defined as the number of instances that correctly and wrongly satisfy rule $r_k$, being $\hat{y}_k=y_j$ and $\hat{y}_k\neq y_j$ respectively; conversely, $TN(r_k)$ and $FN(r_k)$ represent the number of samples $(\x_j, y_j)$ not meeting at least one condition in rule $r_k$, with $\hat{y}_k \neq y_j$ and $\hat{y}_k = y_j$, respectively.

\noindent Then, \emph{rule relevance} $R(r_k)$ of rule $r_k$ can be found as: 
\begin{equation}
    R(r_k)= C(r_k)\cdot ( 1-E(r_k) )
\label{eq:relevanceLLM}
\end{equation}
\subsection{Geometrical Rule Similarity}\label{subsec:overlap}
So far we have shown how we can treat each single rule $r_k$ as a geometrical shape, i.e., the hyper-rectangle $\mathcal{H}_{r_k}$ with volume $\mathcal{V}(\mathcal{H}_{r_k})$, that accounts for all the dimensionalities of the data at hand. 
In this Section, we will describe how such hyper-rectangles are at the basis of a new \textit{geometrical rule similarity} metric, which will be useful to quantify rule overlaps.
Generally speaking (for whatever design choice), rule similarity between two generic rules $r_k$ and $r_z$ of a ruleset $\mathcal{R}$ can be expressed as the following mapping \cite{narteni2022bag}:
$$ q: (r_k, r_z) \in \mathcal{R}\times \mathcal{R}\to \mathbb{R},$$
where higher values of $q(r_k, r_z)$ reflect larger similarity, being $q(r_k, r_z) = 0$ if the rules are completely different. 

As previously mentioned, our aim is designing a rule similarity metrics suitable for quantifying the overlap or the extent of adjacency between two rules, intended as the geometrical intersection (adjacency) between the corresponding hyper-rectangles.\footnote{In case of rule-based models that generate mutually exclusive rules (e.g, a decision trees), there are no overlaps, but still rules can be adjacent, i.e., sharing a smaller or larger surface, which we want to quantify through the proposed metric.}.

\smallskip

Let us denote the two rules with $r_k$ and $r_z$ (both from the same set $\mathcal{R}$).
An overlap (or adjacency) occurs if the following holds:

\begin{equation}
  \max(l_{i_k},l_{i_z}) \leq \min(u_{i_k},u_{i_z})\; \forall i = 1,\ldots,D
\label{eq:overlap}
\end{equation}
%

Assuming that Eq. \ref{eq:overlap} is satisfied for the rules, we can compute the volume of the hyper-rectangle formed by the intersection of the hyper-rectangles $\mathcal{H}_{r_k}$ and $\mathcal{H}_{r_z}$ as follows:
\begin{equation}
    \mathcal{V}_{\text{overlap}(\mathcal{H}_{r_k},\mathcal{H}_{r_z})} = \prod_{i =1}^{D} |\min(u_{i_k},u_{i_z})-\max(l_{i_k},l_{i_z})| 
    \label{eq:volumehyperrect_overlap}
\end{equation}

\smallskip

Finally, Eq. \ref{eq:volumehyperrect} applied to $r_k$ and $r_z$ and Eq. \ref{eq:volumehyperrect_overlap} allow us to define \textit{geometric rule similarity}:

\begin{equation}
    q(r_k, r_z) \doteq \frac{\mathcal{V}_{\text{overlap}(\mathcal{H}_{r_k},\mathcal{H}_{r_z})}}{\mathcal{V}_{\mathcal{H}_{r_k}} + \mathcal{V}_{\mathcal{H}_{r_z}} - \mathcal{V}_{\text{overlap}(\mathcal{H}_{r_k},\mathcal{H}_{r_z})} }
    \label{eq:geometricrulesim}
\end{equation}

\subsection{CONFIDERAI score function}
\label{sec:conformalScore}

Given a rule $r_k$ generated by a rule-based model after training, and predicting an output class $y$, its boundary can be identified as a hyper-rectangle as described in Section \ref{subsec:RBnotation}.
Points lying inside the rule hyper-rectangle, but farther from the boundary are most probably well conforming to the rule output. 
Conversely, if a point is located closer to its boundary, the surroundings of rule $r_k$ need to be considered. Specifically, two situations may occur: \begin{inparaenum}[i)]\item $r_k$ overlaps or is adjacent to other rules predicting the \textit{same class label} $y$ and/or \item $r_k$ overlaps or is adjacent to other rules predicting the \textit{opposite class} (denoted with $\neg y$) \end{inparaenum}. 
Hence, we require our score to penalize more the latter scenario, while not penalizing the first. Geometrical rule similarity helps us designing this behavior. 

\smallskip

Let us now introduce the quantity:
\begin{equation}
\hat{\gamma}(\x, r_k) \doteq  \gamma (\x, r_k) \cdot \frac{\bar{q}(r_k, \mathcal{R}_{\x}^y \backslash r_k )}{\bar{q}(r_k, \mathcal{R}_{\x}^{\neg y})}
\label{eq:hatgamma}
\end{equation}

The first term, $\gamma(\x,r_k)$, encodes the distances of a point $\x$ from the boundary of rule $r_k$ and is computed as:

\begin{equation}
\gamma (\x, r_k) = \sum_{i=1}^{D} \bigg (\frac{1}{d_{i}^{-}(\x,c_{i_k})}+\frac{1}{d_{i}^{+}(\x,c_{i_k})}\bigg ),
\label{eq:gamma}
\end{equation}
with 
\begin{eqnarray*}d_{i}^{-}(\x,c_{i_k}) = |x_i -l_{i_k}|&\text{and} & d_{i}^{+}(\x,c_{i_k}) = |x_i -u_{i_k}| \end{eqnarray*}
and let $\varphi(d):\mathbb{R}\to\mathbb{R}$ be a monotonically decreasing (scalar) function.
In this work, we let $\varphi(d)=\frac{1}{d}$, but other choices are possible. For example, one could set $\varphi(d) = \exp(-\alpha d).$
In this way, a variation on $\alpha$ leads to a variation on the velocity of descent, allowing to control it properly.

The second term of (\ref{eq:hatgamma}) handles the possible overlaps/adjacency with other rules, exploiting Eq. \ref{eq:geometricrulesim}. Specifically, the numerator $\bar{q}(r_k, \mathcal{R}_{\x}^y\backslash r_k)$ expresses the average geometrical rule similarity between $r_k$ and the set $\mathcal{R}_{\x}^y\backslash r_k$ of the other adjacent/overlapped rules predicting the \textit{same} class label of $r_k$. Conversely, the denominator is the average rule similarity between $r_k$ and the set $\mathcal{R}_{\x}^{\neg y}$ of its adjacent/overlapped rules predicting the \textit{opposite} class label.

\smallskip

Finally, to let $\hat{\gamma}(\x,r_k)$ vary in the $[0,1]$ range, we apply the sigmoid function to it, as follows:
\begin{equation}
    \hat{\tau}(\x,r_k) = \frac{1}{1+e^{-\hat{\gamma}(\x, r_k)}}
    \label{eq:tauhat}
\end{equation}

The geometrical term $\hat{\tau}(\x,r_k)$ is then used in combination with rule relevance to define the \emph{score} for point $\x$ and class label $y$:

\begin{equation}
     s(\x,y) \doteq \prod_{r_k\in \mathcal{R}_{\x}^{y}} \hat{\tau}(\x,r_k) (1-R(r_k)),\label{eq:llmCP_score}
\end{equation}

\noindent where the product is on the set $\mathcal{R}_{\x}^{y}$ of rules predicting label $y$ and verified by the input point $\x$.
The contribution of rule relevance is expressed through the term $1-R(r_k)$ (and not directly through $R(r_k)$) in order to keep the score low when classification has better performance, that is when rule relevance is higher.

\subsection{Toy examples}\label{subsec:toyex}
\begin{figure}[!t]
\centering
\hspace*{-1cm}
\subfloat[adjacency - $y=0$]{\includegraphics[width=2.5in]{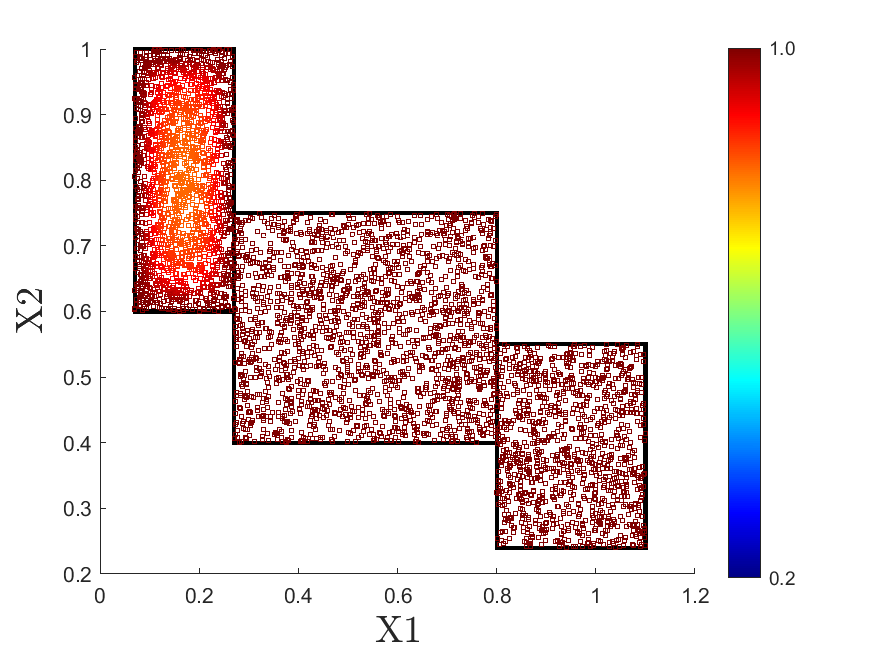}
\label{fig:toyadj0}
}
\hspace{1cm}
\subfloat[adjacency - $y=1$]{\includegraphics[width=2.5in]{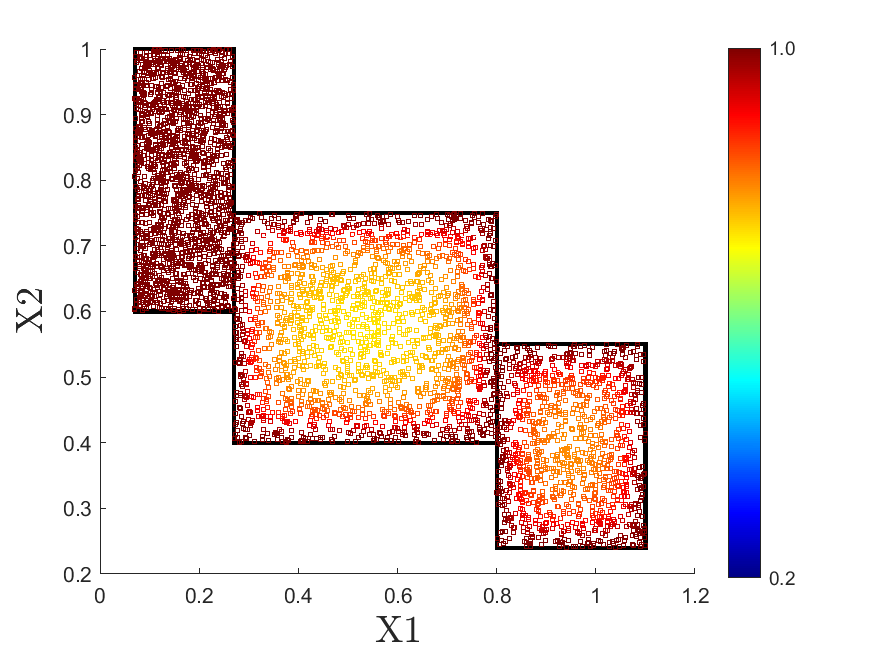}
\label{fig:toyadj1}
}\\
\hspace*{-1cm}
\subfloat[low overlap - $y=0$]{\includegraphics[width=2.5in]{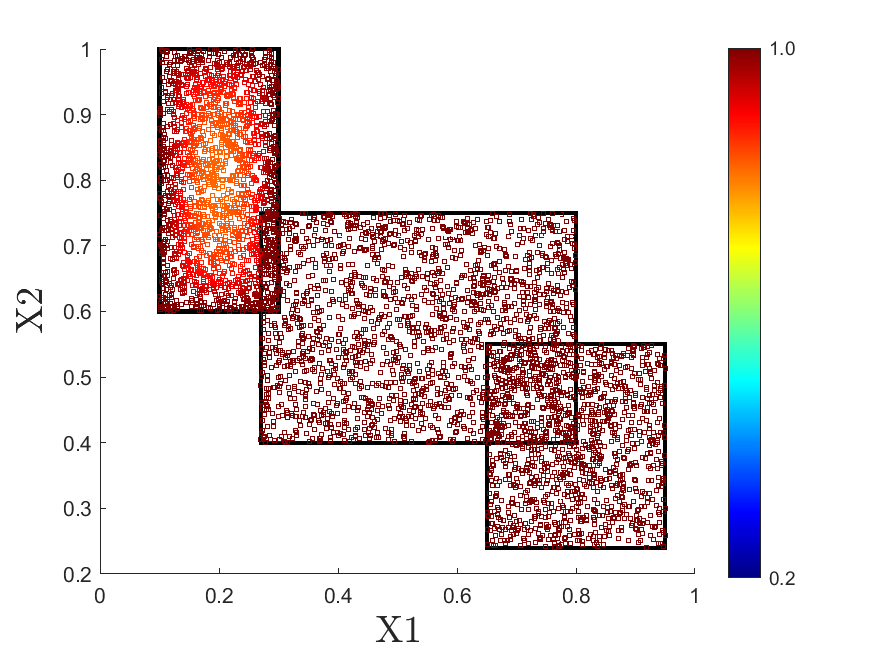}
\label{fig:toymed0}
}
\hspace{1cm}
\subfloat[low overlap - $y=1$]{\includegraphics[width=2.5in]{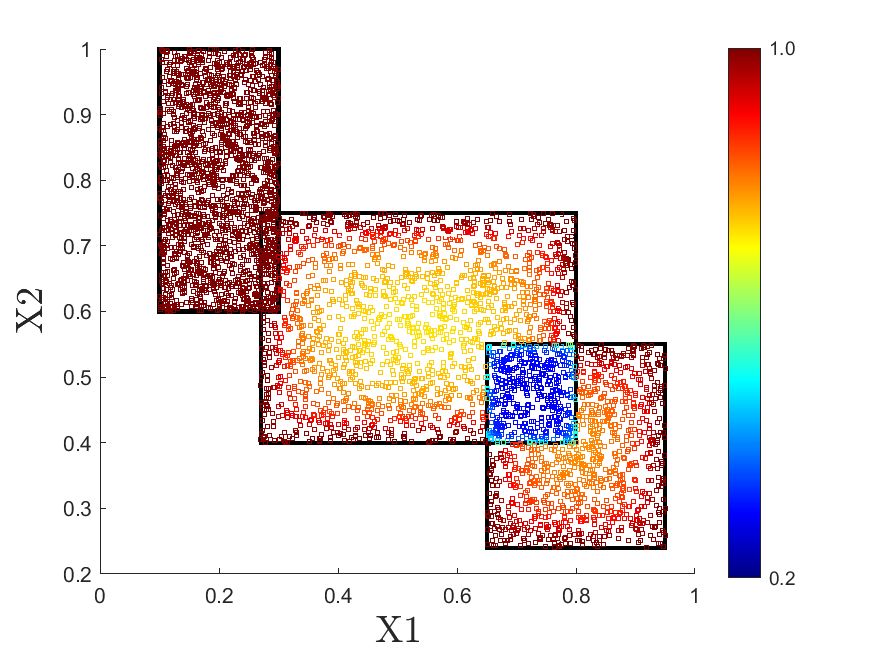}
\label{fig:toymed1}
}\\
\hspace*{-1cm}
\subfloat[high overlap - $y=0$]{\includegraphics[width=2.5in]{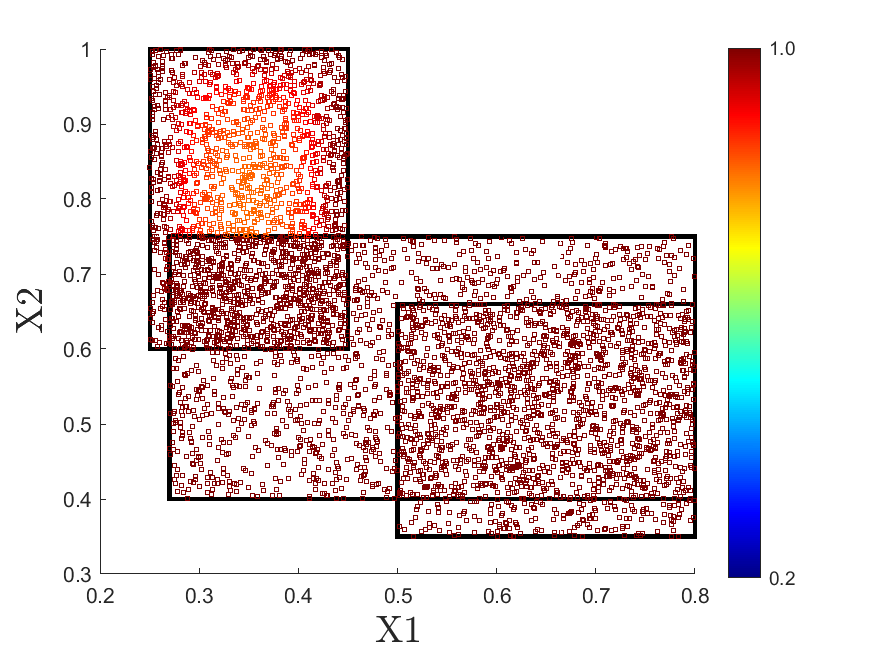}
\label{fig:toyhigh0}
}
\hspace{1cm}
\subfloat[high overlap - $y=1$]{\includegraphics[width=2.5in]{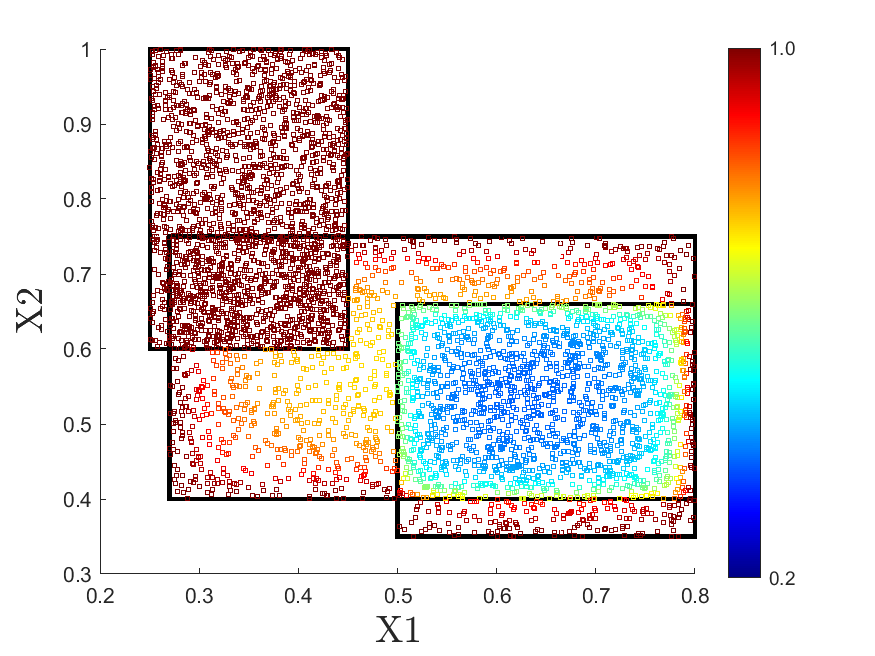}
\label{fig:toyhigh1}
}
\caption{Changes in the values of the score $s(\x,y)$ for points within a set of toy rules designed to have different overlap levels}
\label{fig:newtoyEx}
\end{figure}

Thanks to its design that both accounts for the position of points within the respective hyper-rectangles of rules and incorporates a quantification of rule overlaps via geometrical rule similarity, our score function (\ref{eq:llmCP_score}) is able to capture different degrees of overlap. 

The toy example we present here is devoted to intuitively illustrate how the score is affected when rules with different overlap degrees are considered.
Let us assume a bi-dimensional feature space formed by variables $X_1$ and $X_2$, on which three sets of toy rules are generated. The first set of rules is composed of three rules that are adjacent to each other:

\begin{center}
\footnotesize
$r_1^{\textrm{adj}}$: \textbf{\textcolor{blue}{if}} 0.07 $<$ $X_1$ $\leq$ 0.27 $\land$ 0.6 $<$ $X_2$ $\leq$ 1 \textbf{\textcolor{blue}{then}} $y = 0$\\
$r_2^{\textrm{adj}}$: \textbf{\textcolor{blue}{if}}  0.27 $<$ $X_1$ $<$ 0.8 $\land$ 0.4 $<$ $X_2$ $\leq$ 0.75 \textbf{\textcolor{blue}{then}}  $y = 1$\\
$r_3^{\textrm{adj}}$: \textbf{\textcolor{blue}{if}}  0.8 $<$ $X_1$ $\leq$ 1.1 $\land$ 0.24 $<$ $X_2$ $\leq$ 0.55 \textbf{\textcolor{blue}{then}}  $y = 1$\\
\end{center}
Then, we consider an intermediary case where $r_1^{\textrm{adj}}$ and $r_3^{\textrm{adj}}$ are slightly modified to generate a small overlap with rule $r_2^{\textrm{adj}}$, resulting in the following rules:
\begin{center}
\footnotesize
$r_1^{\textrm{low}}$: \textbf{\textcolor{blue}{if}} 0.1 $<$ $X_1$ $\leq$ 0.3 $\land$ 0.6 $<$ $X_2$ $\leq$ 1 \textbf{\textcolor{blue}{then}} $y = 0$\\
$r_2^{\textrm{low}}$: \textbf{\textcolor{blue}{if}}  0.27 $<$ $X_1$ $<$ 0.8 $\land$ 0.4 $<$ $X_2$ $\leq$ 0.75 \textbf{\textcolor{blue}{then}}  $y = 1$\\
$r_3^{\textrm{low}}$: \textbf{\textcolor{blue}{if}}  0.65 $<$ $X_1$ $\leq$ 0.95 $\land$ 0.24 $<$ $X_2$ $\leq$ 0.55 \textbf{\textcolor{blue}{then}}  $y = 1$\\
\end{center}
Lastly, more pronounced overlaps are considered in the third set: 
\begin{center}
\footnotesize
$r_1^{\textrm{high}}$: \textbf{\textcolor{blue}{if}} 0.1 $<$ $X_1$ $\leq$ 0.3 $\land$ 0.6 $<$ $X_2$ $\leq$ 1 \textbf{\textcolor{blue}{then}} $y = 0$\\
$r_2^{\textrm{high}}$: \textbf{\textcolor{blue}{if}}  0.27 $<$ $X_1$ $<$ 0.8 $\land$ 0.4 $<$ $X_2$ $\leq$ 0.75 \textbf{\textcolor{blue}{then}}  $y = 1$\\
$r_3^{\textrm{high}}$: \textbf{\textcolor{blue}{if}}  0.65 $<$ $X_1$ $\leq$ 0.95 $\land$ 0.24 $<$ $X_2$ $\leq$ 0.55 \textbf{\textcolor{blue}{then}}  $y = 1$\\
\end{center}
These toy rules are useful to understand the behavior of our score function. In this example, we assume that rule relevances are 0, in order to only show the geometrical behavior of the score.
Figure \ref{fig:newtoyEx}, from top to bottom, shows the variation of the score as the overlap between the rules increases. In each plot, rule $r_1^{(\cdot)}$ is the leftmost along the $X_1$ axis, $r_2^{(\cdot)}$ is at the middle, while rules $r_3^{(\cdot)}$ is the rightmost. The left column of the Figure refers to class $y=0$ as the label for computing $s(\x,y)$, and the right column relates to class $y=1$.
A first general observation is that, as expected, the values of the score for  $r_1^{(\cdot)}$ are lower when compute for label $y=0$ than for $y=1$ (where they score 1). Similarly, values for $r_2^{(\cdot)}$ and $r_3^{(\cdot)}$ are lower for label $y=1$ and fixed to 1 for $y=0$. By looking at the plots row-wise, it is possible to see that the score function: 
\begin{inparaenum}[i)] 
\item generates high scores for points lying within the overlap area between rules that predict \textit{different} outputs (see, e.g., the overlap between $r_1^{\textrm{high}}$ and $r_2^{\textrm{high}}$ in Fig. \ref{fig:toyhigh0}, where the score is 1); 
\item generates lower scores for points enclosed in the overlap of rules that predict the \textit{same} output labels (see, e.g., the intersection rectangle between $r_2^{\textrm{high}}$ and $r_3^{\textrm{high}}$ in Fig. \ref{fig:toyhigh1}).
\end{inparaenum}

\section{Logic Learning Machine}
\label{sec:CLLM}

The rule-based model adopted for the experiments of our work is the Logic Learning Machine (LLM), designed and developed by Rulex\footnote{\url{https://www.rulex.ai}} as a more efficient variant of Switching Neural Networks \cite{SNN}. 
Rule generation through LLM takes place in three steps: first, the process starts by discretizing the features and binarizing them via the inverse-only-one coding. The resulting binary strings are then concatenated into a single large string representing the considered samples. Subsequently, shadow clustering is used to build logical structures, called implicants, in the Boolean lattice, which are finally transformed into sets of conditions and combined into a collection of intelligible rules \cite{llmappl1,cangelosi2013logic}.
It is worth underlying that the LLM design process is thus based on an \emph{aggregate-and-separate} approach \cite{mongelli2019performance} able to generate a set of rules that can \emph{overlap}. As a result, an input sample $\x$ may verify multiple rules predicting the same class label and it even may cover rules predicting different output classes.

Let us denote with $\mathcal{R}_{\x}$ the set of all rules satisfied by $\x$. LLM class assignment is then performed based on relevance values.


\noindent Specifically, given a generic point $\x$, 
and the set $\mathcal{R}_{\x}^{y}$ of rules predicting label $y$ and verified by the point, a class label $\hat{y}$ is assigned to $\x$ by solving the following problem \cite{paperEnrico}:
\begin{equation}
    \hat{y} = \argmax_{y} \bigg(\frac{\sum_{r\in \mathcal{R}_{\x}^{y}} R(r)}{\sum_{r\in \mathcal{R}^{y}} R(r)}\bigg)
    \label{eq:llmclass_assignment}
\end{equation}

The topical issue in the conformal framework relies on the fact that the predictions $\hat{y}$ in (\ref{eq:llmclass_assignment}) are not exploited directly as they do not provide any guarantee alone (on the reliability of label assignments). They rather drive the search of guaranteed subspaces of data, on the basis of the set of predictions in $\mathcal{C}_{\varepsilon}(\x)$.

\section{Experimental Results}\label{sec:Ex}
In this Section, we present the results of the experiments devoted to test CONFIDERAI score functions, both in terms of canonical metrics in conformal prediction evaluation (i.e., accuracy, efficiency and computing time, see Sec. \ref{subsec:res_metrics}) and of conformal critical set (Sec. \ref{subsec:res_css}).

All the experiments were executed on a host equipped with Intel Core i5 dual-core processor at 2.6Ghz and 8GB RAM memory. The host runs macOS version 11.7.10.

\subsection{Datasets description}\label{subsec:dataset}
To evaluate the goodness of CONFIDERAI, we tested the method on 10 datasets, which we briefly describe:

\begin{itemize}
    \item \textbf{P2P} and \textbf{SSH}: two datasets concerning peer-to-peer (P2P) and secure shell (SSH) applications of a Domain Name Server (DNS) tunneling detection system \cite{aiello2015dns}; the aim is to detect the presence or absence of DNS attacks by monitoring network traffic and collecting statistical information.
    \item \textbf{BSS}: the Body Signals of Smoking dataset \footnote{Reference link: \url{https://www.kaggle.com/datasets/kukuroo3/body-signal-of-smoking?select=smoking.csv}} collects personal and biological measurements from 
    subjects, with the aim of predicting if these quantities can represent biomarkers of \emph{smoking} or \emph{non-smoking} habits.
    \item \textbf{CHD}: the Cardiovascular Heart Disease dataset\footnote{Reference link: \url{https://www.kaggle.com/datasets/sulianova/cardiovascular-disease-dataset}.} contains patients' records with personal, clinical and behavioral features to predict the presence or the absence of a cardiovascular disease. 
    \item \textbf{Vehicle Platooning}: the dataset consists of simulations of a vehicle platooning system \cite{mongelli} with a binary output of \emph{collision} or \emph{not-collision} under physical features like the number of cars per platoon or the initial distance between cars.
    \item \textbf{RUL}: the Turbofan Engine Degradation Simulation dataset\footnote{Reference link: \url{https://www.kaggle.com/datasets/behrad3d/nasa-cmaps}.} deals with damage propagation modeling for aircraft engines. The goal is to understand which conditions are inherent to imminent faults of the engine by estimating its Remaining Useful Life.
    \item \textbf{EEG}: the Eye State Classification EEG dataset\footnote{Reference link: \url{https://archive.ics.uci.edu/ml/datasets/EEG+Eye+State}.} reports the state of patients' eyes (open or closed) based on continuous electroencephalogram (EEG) measurements.
    \item \textbf{MQTTset} \cite{mqttset}: based on Message Queue Telemetry Transportation communication protocol, this dataset collects measurements from different Internet of Things devices to simulate a smart environment; cyber-attacked data are also included to detect \emph{malicious} and \emph{legitimate} traffic.
    \item \textbf{Magic}: the Magic Gamma Telescope dataset\footnote{Reference link: \url{https://www.kaggle.com/datasets/abhinand05/magic-gamma-telescope-dataset}.} reports Monte Carlo simulations of high energy
    gamma particles in a ground-based atmospheric Cherenkov gamma telescope to distinguish between gamma and hadron radiation.
    \item \textbf{Fire Alarm}: this dataset\footnote{Reference link: \url{https://www.kaggle.com/datasets/deepcontractor/smoke-detection-dataset}.} contains data to develop an AI-based smoke detection device.
\end{itemize}

\subsection{Score function evaluation}\label{subsec:res_metrics}


\begin{table}[htbp]
\begin{adjustwidth}{-2.5cm}{}
  \centering

 \footnotesize
  \caption{Evaluation metrics for CONFIDERAI on Logic Learning Machine model tested on 10 benchmark datasets.}
    \begin{tabular}{cclcccccc}
    \hline
          &Time$_{n_C}$[s] & & \multicolumn{3}{c}{Error} & \multicolumn{3}{c}{Size} \\
    \hline
          &     &  & \multicolumn{1}{c}{\textit{avgErr}} & \multicolumn{1}{c}{\textit{avgErr0}} & \multicolumn{1}{c}{\textit{avgErr1}} & \multicolumn{1}{c}{\textit{avgEmpty}} & \multicolumn{1}{c}{\textit{avgSingle}} & \multicolumn{1}{c}{\textit{avgDouble}} \\
    \hline
    \multirow{4}{*}{\textbf{P2P}} & \multirow{4}{*}{160} & $\varepsilon = 0.01$ & 0     & 0.001 & 0     & 0     & 0.946 & 0.054  \\
\cline{3-9}    &      & $\varepsilon = 0.05$ & 0.039 & 0.079 & 0     & 0.026 & 0.966 & 0.008 \\
\cline{3-9}     &     & $\varepsilon = 0.1$ & 0.048 & 0.079 & 0.018 & 0.035 & 0.965 & 0 \\
\cline{3-9}      &    & $\varepsilon = 0.2$ & 0.152 & 0.226 & 0.077 & 0.152 & 0.848 & 0 \\
    \hline
    \multirow{4}{*}{\textbf{SSH}} & \multirow{4}{*}{200} &  $\varepsilon = 0.01$ & 0.008 & 0.004 & 0.011 & 0     & 0.645 & 0.354  \\
\cline{3-9}     &     & $\varepsilon = 0.05$ & 0.039 & 0.015 & 0.063 & 0.005 & 0.761 & 0.234 \\
\cline{3-9}       &   & $\varepsilon = 0.1$ & 0.094 & 0.05  & 0.139 & 0.039 & 0.822 & 0.139 \\
\cline{3-9}    &      & $\varepsilon = 0.2$ & 0.193 & 0.106 & 0.28  & 0.112 & 0.83  & 0.058 \\
    \hline
    \multirow{4}{*}{\textbf{BSS}} & \multirow{4}{*}{340} & $\varepsilon = 0.01$ & 0.01  & 0.005 & 0.019 & 0     & 0.238 & 0.762  \\
\cline{3-9}       &   & $\varepsilon = 0.05$ & 0.051 & 0.068 & 0.021 & 0.001 & 0.354 & 0.645 \\
\cline{3-9}     &     & $\varepsilon = 0.1$ & 0.101 & 0.127 & 0.053 & 0.008 & 0.484 & 0.508 \\
\cline{3-9}      &    & $\varepsilon = 0.2$ & 0.184 & 0.203 & 0.149 & 0.043 & 0.641 & 0.315 \\
    \hline
    \multirow{4}{*}{\textbf{CHD}} & \multirow{4}{*}{150} & $\varepsilon = 0.01$ & 0.012 & 0.018 & 0.006 & 0.001 & 0.072 & 0.927  \\
\cline{3-9}     &     & $\varepsilon = 0.05$ & 0.049 & 0.057 & 0.042 & 0.001 & 0.239 & 0.76 \\
\cline{3-9}      &    & $\varepsilon = 0.1$ & 0.1   & 0.084 & 0.114 & 0.001 & 0.438 & 0.56 \\
\cline{3-9}      &    & $\varepsilon = 0.2$ & 0.194 & 0.158 & 0.227 & 0.027 & 0.694 & 0.279 \\
    \hline
    \multirow{4}{*}{\textbf{Vehicle Platooning}} & \multirow{4}{*}{133}  & $\varepsilon = 0.01$ & 0.012 & 0.02  & 0.003 & 0     & 0.536 & 0.464 \\
\cline{3-9}    &      & $\varepsilon = 0.05$ & 0.052 & 0.044 & 0.063 & 0.004 & 0.76  & 0.237 \\
\cline{3-9}     &     & $\varepsilon = 0.1$ & 0.102 & 0.099 & 0.104 & 0.034 & 0.844 & 0.122 \\
\cline{3-9}    &      & $\varepsilon = 0.2$ & 0.208 & 0.185 & 0.238 & 0.136 & 0.835 & 0.03 \\
    \hline
    \multirow{4}{*}{\textbf{RUL}} & \multirow{4}{*}{127} & $\varepsilon = 0.01$ & 0.025 & 0.028 & 0.016 & 0.005 & 0.398 & 0.598  \\
\cline{3-9}      &    & $\varepsilon = 0.05$ & 0.058 & 0.06  & 0.051 & 0.007 & 0.557 & 0.436 \\
\cline{3-9}      &    & $\varepsilon = 0.1$ & 0.101 & 0.096 & 0.111 & 0.013 & 0.707 & 0.28 \\
\cline{3-9}       &   & $\varepsilon = 0.2$ & 0.191 & 0.185 & 0.206 & 0.066 & 0.807 & 0.126 \\
    \hline
    \multirow{4}{*}{\textbf{EEG}} & \multirow{4}{*}{300} & $\varepsilon = 0.01$ & 0.02  & 0.015 & 0.026 & 0.001 & 0.327 & 0.672  \\
\cline{3-9}      &    & $\varepsilon = 0.05$ & 0.065 & 0.053 & 0.08  & 0.008 & 0.487 & 0.505 \\
\cline{3-9}     &     & $\varepsilon = 0.1$ & 0.097 & 0.093 & 0.102 & 0.021 & 0.565 & 0.414 \\
\cline{3-9}    &      & $\varepsilon = 0.2$ & 0.19  & 0.194 & 0.185 & 0.083 & 0.67  & 0.247 \\
    \hline
    \multirow{4}{*}{\textbf{MQTTset}} & \multirow{4}{*}{220}  & $\varepsilon = 0.01$ & 0.009 & 0.004 & 0.015 & 0     & 0.705 & 0.295 \\
\cline{3-9}    &      & $\varepsilon = 0.05$ & 0.04  & 0.009 & 0.07  & 0.006 & 0.826 & 0.168 \\
\cline{3-9}     &     & $\varepsilon = 0.1$ & 0.053 & 0.009 & 0.097 & 0.006 & 0.925 & 0.069 \\
\cline{3-9}      &    & $\varepsilon = 0.2$ & 0.167 & 0.091 & 0.24  & 0.144 & 0.794 & 0.062 \\
    \hline
    \multirow{4}{*}{\textbf{Magic}} & \multirow{4}{*}{200} & $\varepsilon = 0.01$ & 0.025 & 0.012 & 0.048 & 0.001 & 0.329 & 0.67   \\
\cline{3-9}       &   & $\varepsilon = 0.05$ & 0.066 & 0.037 & 0.116 & 0.004 & 0.615 & 0.381 \\
\cline{3-9}      &    & $\varepsilon = 0.1$ & 0.13  & 0.116 & 0.155 & 0.039 & 0.688 & 0.273 \\
\cline{3-9}      &    & $\varepsilon = 0.2$ & 0.222 & 0.206 & 0.249 & 0.111 & 0.774 & 0.115 \\
    \hline
    \multirow{4}{*}{\textbf{Fire Alarm}} & \multirow{4}{*}{132} & $\varepsilon = 0.01$ & 0     & 0     & 0     & 0     & 0.973 & 0.027  \\
\cline{3-9}      &    & $\varepsilon = 0.05$ & 0.011 & 0     & 0.022 & 0     & 0.985 & 0.015 \\
\cline{3-9}     &     & $\varepsilon = 0.1$ & 0.077 & 0.133 & 0.022 & 0.077 & 0.91  & 0.013 \\
\cline{3-9}     &     & $\varepsilon = 0.2$ & 0.166 & 0.314 & 0.022 & 0.166 & 0.834 & 0 \\
    \hline
    \end{tabular}%
  \label{tab:metrics}%
  \end{adjustwidth}
\end{table}%

For the evaluation of our score function, we considered canonical accuracy and efficiency metrics for CP for different choices of the error level, namely $\varepsilon = 0.01$, $\varepsilon = 0.05$, $\varepsilon = 0.1$ and $\varepsilon = 0.2$. To provide an indication on the time complexity of the method, we also calculated the time spent for computing the scores on the calibration set, whose size $n_C$ was set to 10000 for all datasets.

Accuracy was measured as the average error, over the test set, of the conformal prediction sets considering points of both classes (\emph{AvgErr}), only class $y=0$ points (\emph{AvgErr0}) and only class $y=1$ points (\emph{AvgErr1}). We remind that an error occurs whenever the true label is not contained in the prediction set.
Efficiency was instead quantified through the rates of test prediction sets with singleton predictions (\emph{Single} for both classes, \emph{Single0} for class $y=0$, \emph{Single1} for class $y=1$), no predictions (\emph{Empty}) and two predictions (\emph{Double}).
The obtained results are reported in Table \ref{tab:metrics}.\\
The overall metrics computed on the benchmark datasets outline the expected behavior of the conformal prediction. 
For all values of $\varepsilon$, the average error is indeed bounded by, and generally linearly increases with, $\varepsilon$ in all cases. In the P2P and Fire Alarm datasets we observe the $avgErr$ is well lower the $\varepsilon$ value, and we can explain this behavior in the relative simplicity of these classification problems. The original LLM models indeed were made up of a small set of rules (6 rules in both cases) and achieved a very high performance (accuracy higher than 98\% for both).
\begin{figure*}[h!]
\hspace*{-3.1cm}
\begin{minipage}[t]{0.5\textwidth}
  \centering
  \includegraphics[width=1.5\linewidth]{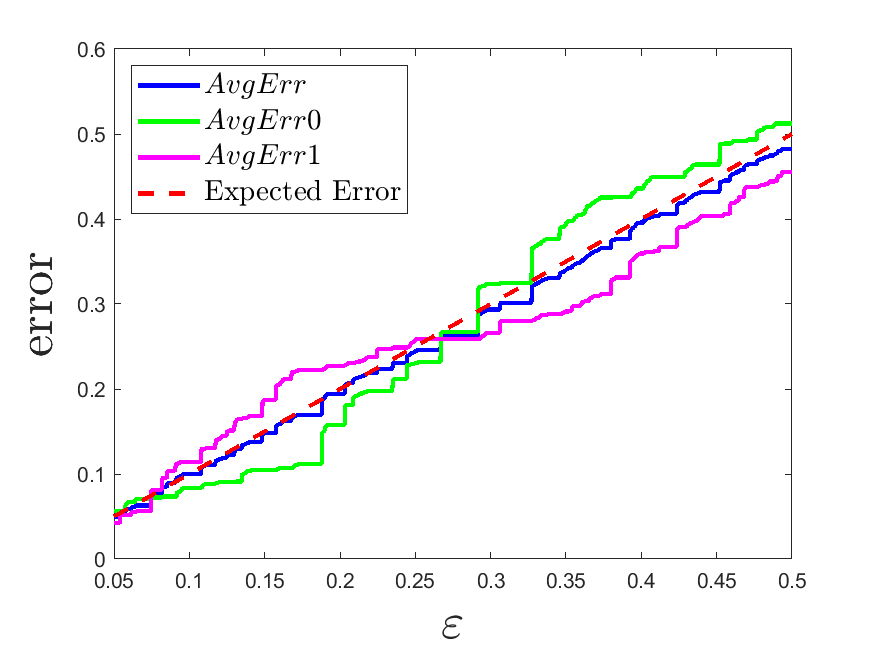}
  \caption*{\hspace*{2.5cm}(\textit{a}) Average errors}
  \label{fig:cardio_error}
\end{minipage}%
\hspace{3.3cm}
\begin{minipage}[t]{0.5\textwidth}
  \centering
  \includegraphics[width=1.5\linewidth]{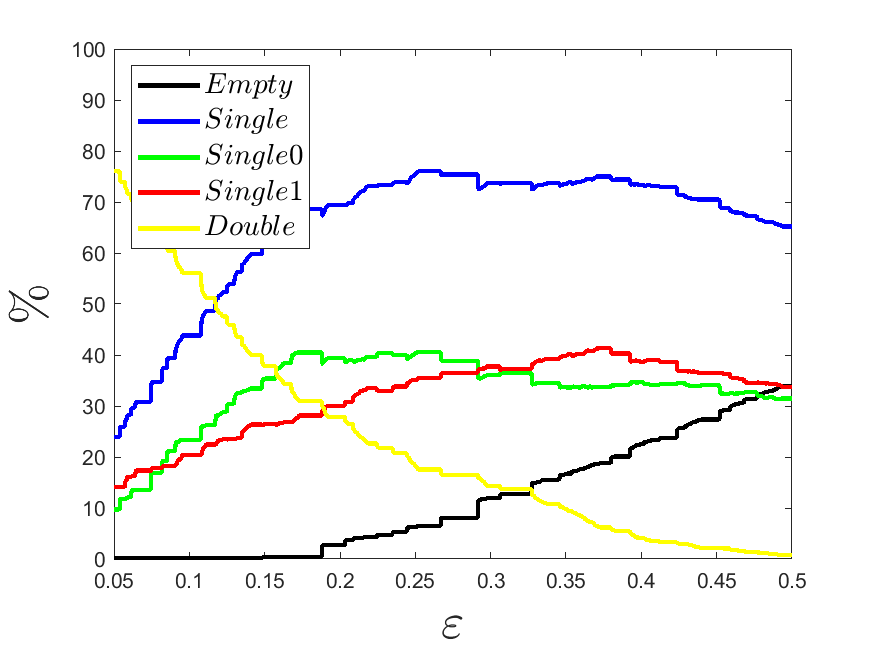}
  \caption*{\hspace*{2.5cm}(\textit{b}) Average sizes}
  \label{fig:cardio_size}
\end{minipage}
\caption{Trend of the performance metrics obtained on the CHD dataset by varying $\varepsilon \in [0.05,\;0.5]$}
\label{fig:plots_cardio}
\end{figure*}
Regarding the size of the conformal prediction sets, we can notice that the empty prediction set rates grow with $\varepsilon$, and that the double-sized regions decrease with it. This is a coherent behavior, since at low $\varepsilon$ the CP algorithm tends to put more labels in the prediction set to maintain the error under that low bound. Conversely, when the error level is allowed to increase, the algorithm reduces the two-labels sets while the singleton predictions increase, and the empty ones increase.
As an example, we illustrate the metrics' trend for the CHD dataset in Fig. \ref{fig:plots_cardio}, for $\varepsilon \in [0,1]$. 
We can observe that the average error on class 0 , i.e. the \emph{healthy} samples, is lower than the average error on class 1, i.e. the \emph{disease} class, up to an $\varepsilon$ value around 0.27; for values larger than this value the trend reverses. 
Concerning the size, we can notice that for about $\varepsilon = 0.12$ the singleton and double-size prediction regions occur in the same percentage (around 50\%), with no empty predictions. For larger $\varepsilon$, singleton predictions rate continues to grow up to about $\varepsilon = 0.33$, where the empty regions increase (with double-sized sets continuing to decrease).

The computation time across the calibration sets is on average 196$\pm$73 s. Its variation is mainly due to two factors: on the one hand, the number $D$ of features of the dataset influences both the computational cost of the geometrical rule similarity ratio and the $\gamma$ terms of Eq. \ref{eq:hatgamma}, resulting in larger times for datasets with more dimensions, such as the BSS for which $D=19$; on the other hand, the number of rules $M_r$ generated by the LLM also influences the computation of the score: the larger is $M_r$, the higher is the chance to have multiple rules covering the same samples, resulting in more terms to be multiplied in Eq. \ref{eq:llmCP_score}. For example, the quickest computation (127 s) was achieved on the RUL dataset which has $D=7$ and $M_r = 42$ rules, while the CHD took a longer time (150 s) having the same $D$ but $M_r = 72$ (i.e., almost 80\% more rules).

\subsection{Evaluation of Conformal Critical Sets}\label{subsec:res_css}

As per Equation \ref{eq:css}, a conformal critical set at a fixed  $\varepsilon$ can be identified. Subsequently, test points belonging to this set can be labelled as \emph{conformal-critical}, providing a new way to look at the dataset.  Specifically, for each point $\x_i$ of the dataset, we can define a new label $\Tilde{y}_i$ as follows:
\begin{equation}
\Tilde{y}_i = \begin{cases} +1 \quad\quad \text{if} \quad \x_i\in \mathcal{S}_\varepsilon,\\
        -1 \quad\quad \text{otherwise}.\end{cases}
        \label{eq:ccslabeling}
\end{equation}
Then, we can train again the LLM model using this labeling to find a new ruleset $\mathcal{R}_{\mathcal{S}_\varepsilon}$, whose rules predicting label $+1$ provide an interpretable description of the CCS.
The identification of new rules to characterize the CCS  boundaries proves very important in real applications, since going outside of them identifies a zone in the feature space where the correct classification of $+1$ points is no more guaranteed, hence other solutions should be sought, such as another training configuration, another model, etc. In light of the Trustworthy AI principle of technical robustness and safety, this result is crucial.
\begin{table}[!h]
  \centering
  \caption{Performance of the LLM trained with new labels from Eq.\ref{eq:ccslabeling}}
    \begin{tabular}{lccc}
    \hline
          & \textbf{TPR} & \textbf{PPV} & \textbf{F1} \\
    \hline
    \textbf{SSH} & 0.64  & 0.94 & 0.76 \\
    \hline
    \textbf{P2P} & 1.00  & 0.93 & 0.96 \\
    \hline
    \textbf{BSS} & 0.55  & 0.62 & 0.58\\
    \hline
    \textbf{CHD} & 0.47  & 0.80 & 0.59\\
    \hline
    \textbf{Vehicle Platooning} & 0.72  & 0.86 & 0.78\\
    \hline
    \textbf{RUL} & 0.53  & 0.69 & 0.60\\
    \hline
    \textbf{EEG} & 0.44  & 0.77 & 0.55\\
    \hline
    \textbf{MQTTSet} & 0.89  & 0.92 & 0.90\\
    \hline
    \textbf{Magic} & 0.63  & 0.90 & 0.73\\\hline
    \textbf{Fire Alarm} & 0.88  & 0.98 & 0.93\\\hline
    \end{tabular}%
  \label{tab:performance_new_rules}%
\end{table}%
For each dataset, we thus trained a new LLM model on the new labels assigned through Equation \ref{eq:ccslabeling} with $\varepsilon=0.05$, getting new rules that were assessed with respect to the ground truth labels, in terms of true positive rate (TPR), precision (also known as PPV) and F1 score. The first two metrics were chosen as the only meaningful performance evaluations of $\mathcal{R}_{\mathcal{S}_\varepsilon}$ are referred to the critical class (i.e., $+1$ labels in our case), which is the target of the CCS. The F1 score measures their balance, being defined as their harmonic mean. Conversely, we disregarded the metrics referred to the other class, since having $\Tilde{y}_i = -1$ does not correspond to predictions for $y=0$, but also for $y=1$ points that do not achieve a singleton prediction region through our score.

Table \ref{tab:performance_new_rules} reports the metrics obtained for all the datasets.
We can observe that the TPR varies among the datasets, from lower values (e.g., EEG or CHD) to better ones (e.g., P2P or platooning), reflecting a different size (here intended as number of points enclosed) of the respective conformal critical sets. Indeed, TPR measures the ratio of correctly predicted $+1$ points with respect to all the points of that class.
On the other hand, PPV values are overall high, exceeding the 80\% in all cases except for BSS, RUL and EEG: this result suggests that the new rules well represent the critical class points, with just a few non-critical target points covered by these rules. Achieving both high TPR and PPV is a known challenge of the precision-recall trade-off, however we managed to get a good balance in the largest part of cases, as pointed out by the high F1 score values (which are computed as the harmonic mean of TPR and PPV).

\smallskip

The metrics reported above refer to point predictions derived, in each dataset, from the different rules of the model, according to Eq. \ref{eq:llmclass_assignment}. Of course, the inspection of the single rules (predicting label $+1$) end up with analogous conclusions. 
\begin{figure}[!h]
\centering
\hspace*{-4cm}
\subfloat[SSH]{\includegraphics[width=0.5\textwidth]{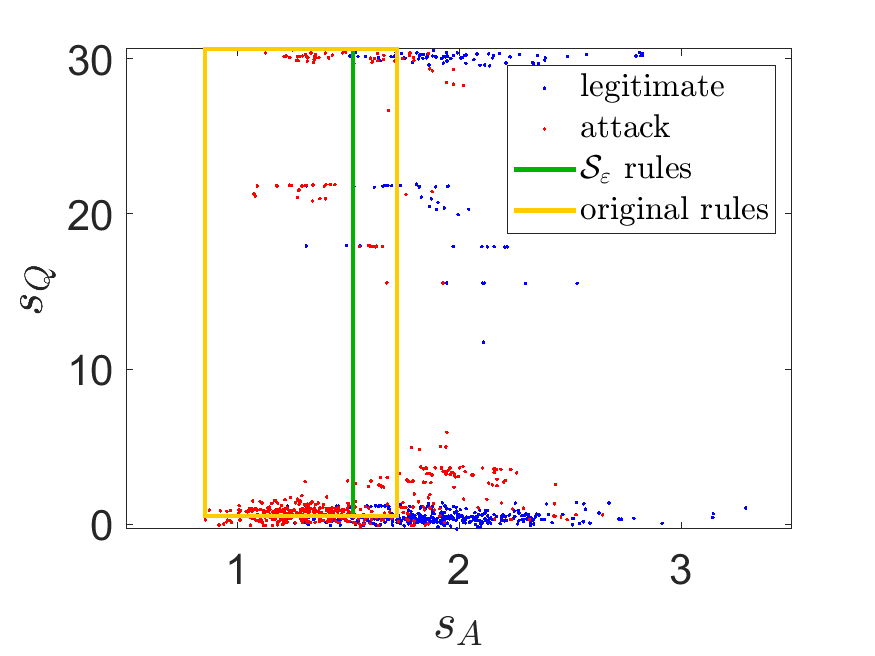}
\label{fig:sshrules}
}
\subfloat[CHD]{\includegraphics[width=0.5\textwidth]{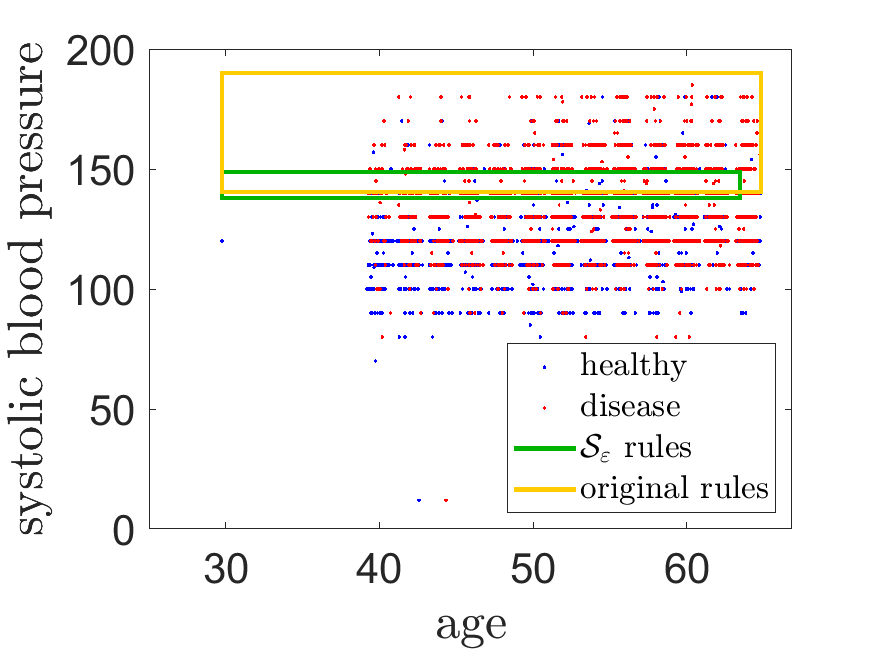}
\label{fig:cardiorules}
}
\subfloat[Vehicle Platooning]{\includegraphics[width=0.5\textwidth]{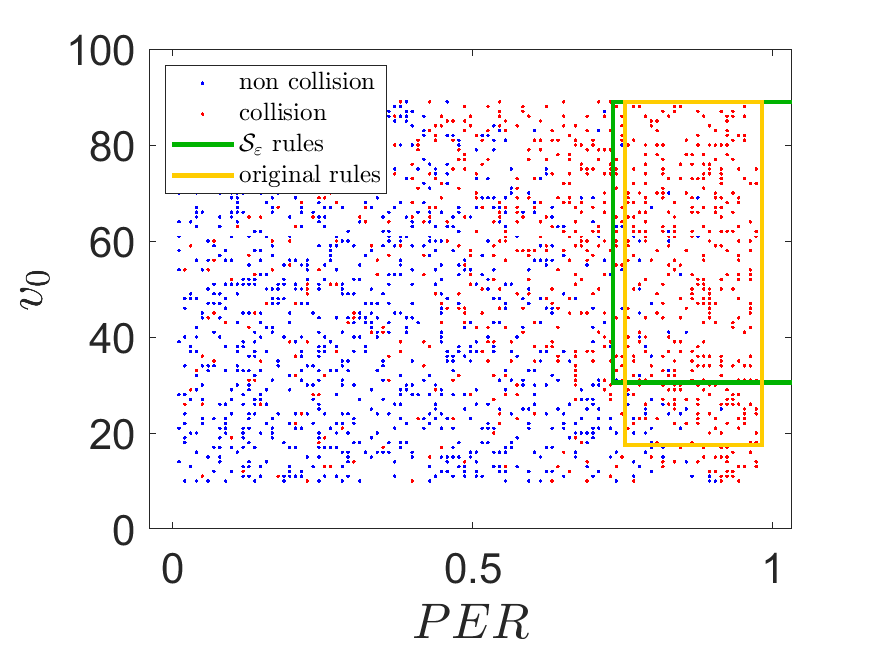}
\label{fig:platrules}
}
\caption{2D scatter plots of three datasets, showing 2D boundaries of the most relevant rules from the original LLM classifier (yellow box) and from the new model derived via the conformal critical set ($\mathcal{S}_\varepsilon$ rule, green box).}
\label{fig:rulescatter_ccs}
\end{figure}
Figure \ref{fig:rulescatter_ccs} shows 2D scatter plots of the classes of the SSH (Fig.\ref{fig:sshrules}), CHD (Fig.\ref{fig:cardiorules}) and platooning (Fig.\ref{fig:platrules}) datasets, where red points denote the critical classes ($y = +1$) we want to characterize and blue points the non-critical ones ($y=0$). We show with a yellow box the first rule by relevance from the original model, and with a green box the top-relevant new rule learned through  $\mathcal{S}_\varepsilon$ (referred to as $\mathcal{S}_\varepsilon$ rule). In all three cases, it can be observed that the new rules leave some blue points out, thus increasing in precision, even if sometimes this dramatically reduces the covering, such as in the CHD case.
However, it is a known trade-off in machine learning between precision and recall: dealing with critical problems, it is often acceptable to have small coverings when reducing the error and increasing precision as much as possible. And this is the behavior exhibited by the examples shown in the figures, whose structure and performance details are reported in Table \ref{tab:ccsrules}. In the SSH dataset, the new most relevant rule brings higher precision, while the error remains the same as in the original rule. In the CHD and vehicle platooning, the rules retrieved by the CCS result in both precision and error improvements with respect to the original case. 
\begin{table}[t!]
\begin{adjustwidth}{-3cm}{-3cm}
  \centering
  \footnotesize
  \caption{Covering, error and precision of the most relevant rule predicting the critical class ($+1$) of the original LLM model and of the new $\mathcal{R}_{\mathcal{S}_\varepsilon}$ model, for the SSH, CHD and vehicle platooning datasets.}
    \begin{tabular}{lllccc}
    \hline
          &       & \multicolumn{1}{r}{} & \multicolumn{1}{l}{\textbf{Covering}} & \multicolumn{1}{l}{\textbf{Precision}} & \multicolumn{1}{l}{\textbf{Error}} \\
    \hline
    \multirow{2}[4]{*}{\textbf{SSH}} & $\mathcal{S}_\varepsilon$ rule & \textcolor{blue}{\textbf{if}} $s_A \leq 1.52 \land s_Q > 0.56 $ \textcolor{blue}{\textbf{then}}  \textit{attack}  & 0.45  & 0.95  & 0.03 \\
\cline{2-6}          & Original rule & \makecell[tl]{\textcolor{blue}{\textbf{if}} $ v_A  \leq 38058 \land v_Q \leq 2095 \;\land $\\$s_A \leq 1.72 \land s_Q > 0.55 $ \textcolor{blue}{\textbf{then}} \textit{attack}} & 0.35  & 0.91  & 0.03 \\
    \hline
    \multirow{2}[4]{*}{\textbf{CHD}} & $\mathcal{S}_\varepsilon$ rule & \makecell[tl]{\textcolor{blue}{\textbf{if}} $age \leq 63\;\land$ $height > 152 \; \land$ $ weight \leq 87 \; \land $ \\ $139 < systolic\; blood \;pressure \leq 149 \;\land$\\ $diastolic\; blood \;pressure > 79 \;\land$ \\ $cholesterol \leq 2.5 \;\land$ $gluc \leq 2.5 $ \textcolor{blue}{\textbf{then}} \textit{disease} }& 0.13  & 0.89  & 0.02 \\
\cline{2-6}          & Original rule & \textcolor{blue}{\textbf{if}} $systolic\; blood\; pressure > 140 $ \textcolor{blue}{\textbf{then}} \textit{disease}  & 0.22  & 0.82  & 0.05 \\
    \hline
    \multirow{2}[4]{*}{\textbf{Vehicle platooning}} & $\mathcal{S}_\varepsilon$ rule & \textcolor{blue}{\textbf{if}} $PER > 0.74 \land v0 > 30$ \textcolor{blue}{\textbf{then}} \textit{collision} & 0.37  & 0.88  & 0.05 \\
\cline{2-6}          & Original rule & \textcolor{blue}{\textbf{if}} $PER > 0.76 \land v0 > 17$ \textcolor{blue}{\textbf{then}} \textit{collision} & 0.44  & 0.86  & 0.07 \\
    \hline
    \end{tabular}%
  \label{tab:ccsrules}%
  \end{adjustwidth}%
\end{table}%
These examples also reveal different possibilities that can occur when moving from the original ruleset to the new one. 
In some cases, the new rules may exclude the role of some features that previously had one, and this kind of guidance helps filtering out variables of the problem at hand that have no impact on the conformal guarantees: it is the case of the SSH, where features $v_A$ and $v_Q$ are no more present in the premise of $\mathcal{S}_\varepsilon$ rule.  
In some other cases, such as the platooning in our example, the conditions remain exactly on the same variables, but with changes to the related thresholds. Finally, it can happen that $\mathcal{S}_\varepsilon$ rules also reveal new factors that were not highlighted in the original rules, as for in the CHD example, where we can observe that the $\mathcal{S}_\varepsilon$ rule expresses much more conditions (on different features) than the original one: in a real clinical application, this might serve to clinicians as a more thorough guidance, revealing all the factors most probably involved in the presence of a cardiovascular disease.
These are therefore short yet important examples motivating the introduction of conformal critical sets to shed light into how rule-based classifiers can be tuned via conformal prediction guarantees to achieve rules with higher guarantees on a target (critical) class.

\section{Conclusion}
\label{sec:conclusion}

This paper introduced CONFIDERAI, a new score function for rule-based models directly designed on top of the properties of these models. Indeed, starting from the decision rules generated by the model, conformity is derived as a function of the placement of the samples with respect to the geometry of the model, also taking into account rule relevance, a measure that reflects the predictive quality of a rule. Moreover, the score takes into account possible rule overlaps thanks to a geometrical rule similarity term. 

Extensive experimentation by considering the Logic Learning Machine model on several datasets has shown a behavior in line with conformal prediction framework, both in terms of accuracy and efficiency of the prediction sets.  
In addition, by leveraging on the results of CONFIDERAI, we moved a step beyond the probabilistic guarantees provided by the conformal predictions, in the direction of a more safety-preserving solution. Thus, we first defined the notion of conformal critical set  that provides guarantee to efficiently predict the target class points (i.e., the critical ones) in high probability (thanks to the CP); then, we retrained the rule-based model on the data newly labelled according to the CCS, and ended up in rules with improved precision and reduced false positives on the target class.

The current work is a starting point for the development of a fully conformal rule-based methodology for trustworthy AI. Future works will involve a more in-depth experimentation of other rule-based models and their assessment on real world-applications, as well as the extension of the proposed score function to multi-class problems. Moreover, further investigation could be devoted to better formalize the transition from the original rules to those arising from conformal critical sets.

\section*{Acknowledgments}

The authors would like to thank Teodoro Alamo of  University of Seville for thoughtful discussions about conformal predictions and probabilistic safety sets and Enrico Cambiaso for inspiring the ideas on geometrical rule similarity.

This work was partially supported by REXASI-PRO H-EU project, call HORIZON-CL4-2021-HUMAN-01-01, Grant agreement ID: 101070028. The work was also supported by Future Artificial Intelligence Research (FAIR) project, Italian Recovery and Resilience Plan (PNRR), Spoke 3 - Resilient AI.

\bibliographystyle{elsarticle-num} 
\bibliography{main}

\end{document}